\documentclass[conference]{IEEEtran}
\IEEEoverridecommandlockouts

\usepackage{mwe}
\usepackage{caption}
\usepackage{subcaption}

\makeatletter
\def\ps@IEEEtitlepagestyle{%
  \def\@oddfoot{\mycopyrightnotice}%
  \def\@oddhead{\hbox{}\@IEEEheaderstyle\leftmark\hfil\thepage}\relax
  \def\@evenhead{\@IEEEheaderstyle\thepage\hfil\leftmark\hbox{}}\relax
  \def\@evenfoot{}%
}
\def\mycopyrightnotice{%
  \begin{minipage}{\textwidth}
  \centering \scriptsize
  Copyright~\copyright~2024 IEEE. Personal use of this material is permitted. Permission from IEEE must be obtained for all other uses, in any current or future media, including\\reprinting/republishing this material for advertising or promotional purposes, creating new collective works, for resale or redistribution to servers or lists, or reuse of any copyrighted component of this work in other works by sending a request to pubs-permissions@ieee.org.
  \end{minipage}
}
\makeatother

\usepackage{multicol}
\usepackage{graphicx}
\usepackage[utf8]{inputenc} 
\usepackage[T1]{fontenc}    
\usepackage{hyperref}       
\usepackage{url}            
\usepackage{booktabs}       
\usepackage{amsfonts}       
\usepackage{nicefrac}       
\usepackage{microtype}      
\usepackage{xcolor}         

\usepackage{graphicx}

\usepackage{amsmath,amssymb,amsfonts}

\usepackage[ruled, lined, linesnumbered, commentsnumbered, longend]{algorithm2e}

\SetCommentSty{mycommfont}

\usepackage{amsmath}
\usepackage{wrapfig}

\usepackage{setspace, etoolbox}
\usepackage{hyperref}
\usepackage{subcaption}

\usepackage{amssymb}
\newcommand\F{\mathcal{F}}

\newcommand\M{\mathcal{M}}

\newcommand\R{\mathcal{R}}
\newcommand\SN{\mathcal{S}}
\newcommand\V{\mathcal{V}}

\newcommand\D{\mathcal{D}}

\newcommand\X{\mathcal{X}}

\DeclareMathOperator*{\argmax}{arg\,max}

\usepackage{cite}
\usepackage{amsmath,amssymb,amsfonts}
\usepackage{graphicx}
\usepackage{textcomp}
\usepackage{xcolor}
\def\BibTeX{{\rm B\kern-.05em{\sc i\kern-.025em b}\kern-.08em
    T\kern-.1667em\lower.7ex\hbox{E}\kern-.125emX}}

\newcommand{\citet}[1]{\citeauthor{#1}~(\citeyear{#1})}
\newcommand{\citep}[1]{\cite{#1}}

\begin{document}

\captionsetup[figure]{labelfont={bf},labelformat={default},labelsep=period,name={Figure}}

\title{Towards Interpreting Multi-Objective Feature Associations}

\author{\IEEEauthorblockN{1\textsuperscript{st} Nisha Pillai}
\IEEEauthorblockA{\textit{Computer Science and Engineering} \\
\textit{Mississippi State University}\\
Starkville, MS, USA \\
pillai@cse.msstate.edu}
\and
\IEEEauthorblockN{2\textsuperscript{nd} Ganga Gireesan}
\IEEEauthorblockA{\textit{College of Veterinary Medicine} \\
\textit{Mississippi State University}\\
Starkville, MS, USA \\
gg733@msstate.edu}
\and
\IEEEauthorblockN{3\textsuperscript{rd} Michael J. Rothrock Jr.}
\IEEEauthorblockA{\textit{Egg and Poultry Production Safety Research Unit} \\
\textit{USDA ARS}\\
USA \\
michael.rothrock@usda.gov}
\and
\IEEEauthorblockN{4\textsuperscript{th} Bindu Nanduri}
\IEEEauthorblockA{\textit{College of Veterinary Medicine} \\
\textit{Mississippi State University}\\
Starkville, MS, USA \\
nanduribindu@gmail.com}
\and
\IEEEauthorblockN{5\textsuperscript{th} Zhiqian Chen}
\IEEEauthorblockA{\textit{Computer Science and Engineering} \\
\textit{Mississippi State University}\\
Starkville, MS, USA \\
zchen@cse.msstate.edu}
\and
\IEEEauthorblockN{6\textsuperscript{th} Mahalingam Ramkumar}
\IEEEauthorblockA{\textit{Computer Science and Engineering} \\
\textit{Mississippi State University}\\
Starkville, MS, USA \\
ramkumar@cse.msstate.edu}
}

\maketitle

\begin{abstract}

Understanding how multiple features are associated and contribute to a specific objective is as important as understanding how each feature contributes to a particular outcome. Interpretability of a single feature in a prediction may be handled in multiple ways; however, in a multi-objective prediction, it is difficult to obtain interpretability of a combination of feature values. To address this issue, we propose an objective specific feature interaction design using multi-labels to find the optimal combination of features in agricultural settings. One of the novel aspects of this design is the identification of a method that integrates feature explanations with global sensitivity analysis in order to ensure combinatorial optimization in multi-objective settings. We have demonstrated in our preliminary experiments that an approximate combination of feature values can be found to achieve the desired outcome using two agricultural datasets: one with pre-harvest poultry farm practices for multi-drug resistance presence, and one with post-harvest poultry farm practices for food-borne pathogens. In our combinatorial optimization approach, all three pathogens are taken into consideration simultaneously to account for the interaction between conditions that favor different types of pathogen growth. These results indicate that explanation-based approaches are capable of identifying combinations of features that reduce pathogen presence in fewer iterations than a baseline. 


\end{abstract}

\section{Introduction}

Imagine a farmer wanting to reduce the risk of food-borne illness associated with their farming practices. Their agricultural objectives might have been achieved by combining several practices with a variety of parameters. Often, it is difficult to identify a combination of practices and specific parameters that do not exacerbate the presence of food-borne pathogens. The solutions to such combinatorial optimizations can currently be obtained using either exact solutions (dynamic programming)~\cite{xu2020deep, Pillai_CO2023} that guarantee an optimal solution, or approximate methods (genetic algorithms, Bayesian optimization)~\cite{weissteiner2023bayesian, acampora2023genetic}. In contrast to exact solutions, approximation methods are computationally efficient but offer a lower level of accuracy. The best approach here is to strike a balance between exact and approximate methods.

\begin{figure*}[t]
\centering
\includegraphics[width=0.98\textwidth]{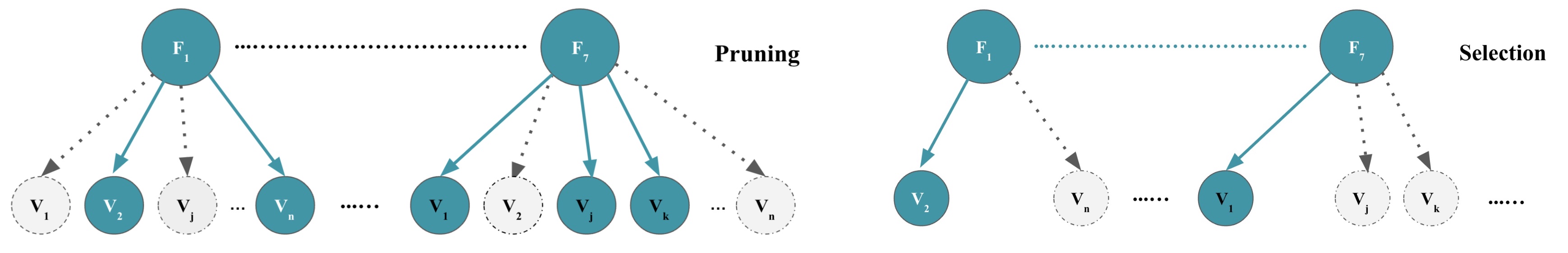}
\caption{Selection of the best feature value combinations based on the importance of the features and objective criteria. To reduce the combinatorial optimization search space, we use popular explanation methods to find local explanations of each feature (root node) and value (child node). A threshold on feature relevance is used to prune combinations of features and values from the search space. The final combination selection is based upon the objective of the problem and the relevance of the feature combinations to the model prediction. }
\label{fig:pruning}
\end{figure*}

As an alternative to approaching combinatorial optimization (CO) as a traveling salesman problem~\cite{yang2023memory}, we propose adding the importance of the features to meet a special goal as a weight to the problem of combinational optimization. This study integrates explainable AI~\cite{gunning2019xai} and combinatorial optimization to arrive at an optimal solution in a multi-objective problem. To attain the desired combinations earlier in the process, we add feature importance-based pruning and objective-based selection during the selection of the best feature value combinations (Fig. \ref{fig:pruning}). In this research, a key contribution is to ensure an optimal multi-objective feature value selection approach by combining additive features attribution methods with selective pruning  that are applicable to agricultural settings in order to reduce food-borne diseases.

The outline of this paper is as follows. Related research is included in Section~\ref{related}. The proposed approach is presented in Section~\ref{approach}. Section~\ref{dataset} examines two multi-objective datasets used in this research to evaluate the problem. The experiments and results are described in Section~\ref{experiments}. Conclusions are in Section~\ref{conclusion}.

\section{Related Research} \label{related}

In real life, combinatorial optimization (CO) problems are encountered in a wide variety of problem groups. The complexities of these problems make them challenging to resolve~\cite{kaya2022review}. Historically, researchers have attempted to resolve these problem groups~\cite{weinand2022research} including the quadratic assignment problem~\cite{hurtado2022exact}, the minimum spanning tree problem~\cite{majumder2022multi}, the location-routing problem~\cite{tahami2022literature}, and the travelling salesman~\cite{panwar2023discrete} problem using exact or approximate solutions. Exact approaches popularly available to solve CO algorithms, includes branch-and-bound~\cite{zhang2022deep} and integer programming~\cite{zhang2023survey}. NP-hard combinatorial problems are difficult to solve in many instances, and exact solutions are not always feasible.  Metaheuristic algorithms such as evolutionary algorithms (EAs)~\cite{de2022random}, such as genetic algorithms (GA), are useful in addressing NP-hard problems (CO)~\cite{zhao2022evolution}. Physics-based algorithms~\cite{geist2023combinatorial}, swarm-based algorithms~\cite{zhu2022improved}, bio-inspired algorithms, and nature-inspired algorithms utilize feedback based approaches in order to find suboptimal solutions to such NP-hard problems. This research examines the application of feature explanations to obtain the optimal combinatorial solution for a multi-objective problem.

Generally, explainable AI algorithms~\cite{laato2022explain} can be divided into four categories: 1) explanation by simplification, 2) explanation by feature relevance, 3) local explanation, and 4) explanation by visuality. The use of explainable methods that ensure transparency has a number of applications~\cite{panigutti2022understanding, ahmed2022artificial} and we utilize the feature relevance explanation in achieving reduced food-borne pathogens in agricultural systems.

A deep SHapley additive explanation (DeepSHAP)~\cite{lundberg2017unified} methodology is used in this research to provide reliable results and to explain the contribution of each agricultural practice to the food-borne pathogen prediction model. SHAP based explanations focus primarily on local explanations of each feature, while our combinatorial problem requires a global measure of the combination of features. In our application, we focus on finding the correlated features that will help in a multi-objective problem. To reduce the search space of <feature, value> combinations, we carry out a variance-based global sensitity analysis.  

\section{Explainable AI for Combinatorial Optimization} \label{approach}

\begin{figure*}[t]
\centering
\includegraphics[width=1\textwidth, trim={0 2.1cm 0.05cm 3.55cm},clip]{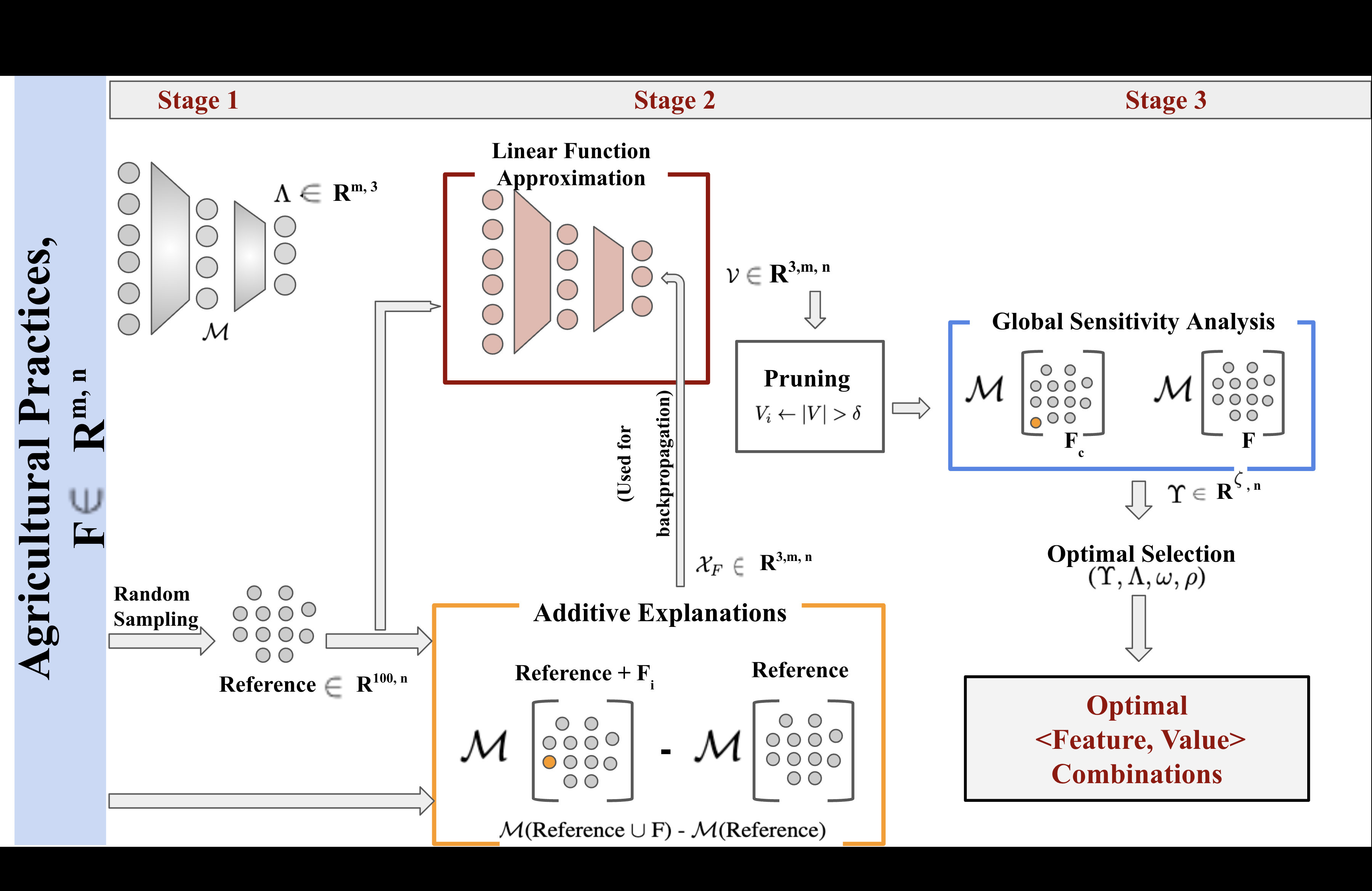}
\caption{Design diagram of multi-label combinatorial optimization. To predict food-borne pathogens in the first stage of the process, we build a multi-label multi-layer perceptron architecture based on our agricultural practices. The next step involves learning the relevance of a feature, value> relevance in the prediction using DeepShap~\cite{lundberg2017unified} which is combined with additive explanations (SHAP) and linear functional approximations (DeepLIFT). By using threshold-based pruning, the search space is optimized and computing time is reduced. In the final step of our study, our objective based selection is based on learning the global explanation of feature combinations.   }
\label{fig:design}
\end{figure*}

This section presents a combinatorial optimization framework (Fig. \ref{fig:pruning}) for finding optimal feature value combinations in multi-label settings. This framework (See Fig. \ref{fig:design}) consists of five steps: calculating feature importance using additive explanations, reducing computation time through linear compositional approximation, pruning by threshold to reduce problem search space, global sensitivity analysis to determine feature combination importance, and optimal feature selection. Algorithm~\ref{alg:featurelearn} provides a detailed description of the algorithm.

\subsection{Problem Statement}
This research uses dataset $\D$, which consists of input data $\F \in R^{ m, n}$ and output variables $Y \in R^{ m, 3}$. This study uses $m$ samples and $n$ features to learn a multi-objective (3, with our dataset) classification algorithm.  In this problem, we are interested in finding the optimal combination of features ($\F$) and values ($\V$)  that will meet a specific objective. Towards discovering optimized combinations, we build a single hidden layer neural network for multi-label problems.

\subsection{Multi-Label Neural Network Architecture} 
We use a single hidden layer neural network to simplify the architecture to predict $Y$ based on input variables $\F$. Our approach for multi-label classification uses sigmoid non-linear activation at the last layer, and for multi-label regression, we use a linear activation function. We use our trained model $\M$, to determine the importance of features and values for the given problem.

\subsection{Additive Feature Explanations}

A contribution of each feature and value is necessary to meet the objective of finding the optimal combination of features and values. As a first step in extracting the relevance score of feature value combinations, we use SHapley Additive Explanations (SHAP)~\cite{NIPS2017_7062}, which ensure efficiency, consistency, and symmetry. An explanation of a feature in prediction is derived by estimating marginal contributions based on a comparison between prediction scores of a reference subset that includes and excludes the feature. Formally, given a random Reference data $\R$, the Shapley value for feature $F_i$ is computed as the following:
\begin{equation}
\X(\R)_{\F i}  \gets \dfrac{1}{d} \sum_{\R  \subseteq \F \setminus \F_{i}} \begin{bmatrix}
        d - 1 \\
        |\R|
    \end{bmatrix}^{-1}
 \M (\R \cup \F_i) - \M (\R)
\end{equation}
By using this approach, we can calculate the explanations of each feature and value in the network prediction.

\subsection{Linear Compositional Approximation with Deep Networks}

\SetKwComment{Comment}{$\triangleright$\ }{}
\SetKwInput{KwDefinition}{Definition}

\begin{algorithm}[t]
\DontPrintSemicolon
\SetAlgoLined
\SetNoFillComment
\setstretch{1.25}
\KwDefinition{ EBCO (learned multi-label model $\M$, dataset $\F$, pruning threshold $\delta$, prediction significance $\omega$, sensitivity threshold $\rho$, long-term memory cut $\zeta$)}

\KwResult{Feature combinations with highest importance and lowest multi-pathogen presence $\SN$ } 
$\ $
$\SN \gets \{\}$\\
$\Lambda \gets $ model.predict($\M$, $F$)\\
Reference $ \gets $ RandomSampling($F$)\\
$\Psi  \gets  {\F_0}$ \\
\While{$ \exists \F_i \in \Psi$} {

$\ $$\X_{F}\gets $ AdditiveExplanations ($\M$, Reference, $\F_i$)\\
$ \V \gets $ LinearFunctionApproximation($\X_{F}, \F_i$)\\ 
$ \V_{i} \gets $Multi-LabelPruning ($\V, \delta$)\\
\For{$v\ in\ \V$}
{  
$\ $$\Upsilon \gets $ GlobalSensitivityScore($v$, $\rho$)\\
$\Gamma \gets$ OptimalSelection ($\Upsilon, \Lambda, \omega, \rho$)\\
$\SN \gets \argmax_{} \Gamma_{f, v} $
}
$\SN \gets$ select  $\zeta$ combinations\\
$\Psi  \gets  {\F_{i+1}}$ 
}

\caption{Explanation based multi-label combinatorial optimization}\label{alg:featurelearn}
\end{algorithm}

Additive eXplanations provide a simplified definition of the measure of feature importance. Calculating the exact computation time for each feature, value importance is however challenging. To approximate the computation of these expected values, deep neural networks can be advantageously utilized due to their compositional nature.  With DeepLIFT~\cite{shrikumar2016not, shrikumar2017learning}, it is possible to estimate the feature importance by using neural networks based on the difference between two reference sets with and without a particular value in the backpropagation method. With the combined architecture of SHapley Additive Explanations and DeepLIFT, the SHAP value computations can be approximated and ensured to be accurate and consistent. The DeepSHAP~\cite{lundberg2017unified} method combines SHAP values computed for smaller components of the network into SHAP values for the entire network utilizing the DeepLIFT method. A rough estimate of each feature's significance ($\V$) is obtained by recursively passing SHAP's computed feature importance as the loss function of DeepLIFT through the network in backpropagation.

\subsection{Threshold Based Pruning}

Linear Approximation provides the interpretation of individual features, however calculating interactions between all these feature values would be a highly complex and time-consuming computation. Using a threshold ($\delta$) that meets our multi-objective requirements, we prune the feature-value combination tree to reduce its complexity. 
\begin{equation}
V_i \gets |V| > \delta 
\end{equation}

\subsection{Global Sensitivity Analysis}

The significance of features is determined using either statistical variance tests or a regular regression framework, or by analyzing the weight of a neural network. Functional decomposition applied to variance can also be used to compute the importance of individual features or interactions between features.  A popular approach, global Sensitivity Analysis evaluates the importance of an input variable $F_i$ by how much variance $F_i$ contributes to $Y,$ so if we condition $F_i$, we look at how much variance has been reduced. The uncertainty associated with $Y$ can be attributed to the uncertainty associated with $F_i$ since it represents much of its variance. A variance-based global sensitivity analysis~\cite{azzini2021sobol} is employed in this research to determine the influence of feature associations in predictions. Our first step is to fix the associated features ($F_q$) in question in the original feature set $F$ to create a new feature set $F_c$.  
\begin{equation}
F_c \gets F_1......F_k, F_{q1}, F_{q2}.....F_n
\end{equation}

For finding the relevance or variance of $Y$ given the new feature set, we calculate the covariance between $E(\M(F_c|Y))$ and $E(\M(F|Y)$. We also use variance in $Y$ as a normalizing parameter, Variance($F|Y$)
\begin{equation}
\Upsilon \gets \dfrac{COV(E(\M(F_c|Y)), E(\M(F|Y)))}{VAR(E(\M(F|Y)))}
\end{equation}

\subsection{Optimal Selection}

To satisfy our objective of finding features that can meet our expectations, we use prediction-based selection at this stage. If our objective is to find combinations that reduce the prediction, we should use $1 - \Lambda$ in the expression. Our objective is achieved by adding a prediction significance threshold, $\omega$ part of prediction to a 1 - $\omega$ part of feature importance, $\Upsilon$. 
\begin{equation}
\gamma_i \gets \omega * (1 - \Lambda_i) + (1 - \omega) * \Upsilon_i
\end{equation}

In the multi-objective scenario, we penalize low $\gamma$ values with a threshold $\rho$, to 0 and add to get a single definitive score that is used to calculate feature association relevance. 
\begin{equation}
\Gamma_i \gets \sum_{o \in \gamma} \begin{cases}
      0, & \text{if}\ o < \rho \\
      \rho, & \text{otherwise}
    \end{cases}
    \label{eq:relevance}
\end{equation}

Following this, a selection of feature value combinations ($\SN$) is made to provide explanations that meet the highest expectations. Each feature is ranked sequentially, and its score is calculated for every selected value in equation \ref{eq:relevance}. The combinations tree is reduced at every stage with a long-term cut, $\zeta$, which ensures a reduction in computation. 
\begin{equation}
\SN \gets \zeta\  combinations\  \argmax_{} \Gamma_{f, v} 
\end{equation}

\section{Dataset} \label{dataset}
The results of two multi-label problems are presented in this paper. We intend to reduce the multi-drug resistance of three pathogens associated with pre-harvest poultry management practices by using the first dataset. Our objective in the second problem is also a multi-label classification, where we are trying to determine the most optimal combination of post-harvest poultry practices which will result in the absence of pathogens. 

\subsection{Multi-drug resistance reduction in pre-harvest pastured poultry practices}

The data used in this study is pastured poultry dataset~\cite{hwang2020farm, pillai2022ensemble} that was collected from eleven pastured poultry farms located in the southeastern United States over a period of four years and is described in detail previously~\cite{ayoola2022preharvest}. Pre-harvest samples (feces and soil) were taken at three timestamps: within a few days after broilers were placed on pasture, halfway through their pasture stay, and on the day the flock was processed. A minimum of 25 grams of sample was collected for each field sample. Using phosphate-buffered saline (PBS) diluted 1:3 with three grams (feces, soil) was combined within filtered stomacher bags (Seward Laboratories Systems, Inc., West Sussex, UK). All samples were homogenized for 60 seconds, and homogenates were used for all downstream cultural isolation of \textit{Salmonella}, \textit{Listeria}, and \textit{Campylobacter}. The published NARMS protocols and NARMS breakpoints were used for characterization the antibiotic resistance for all three pathogens and isolates that were resistant to three or more tested antibiotics were considered to be multidrug resistant (MDR). We examined general poultry management settings (31 farm and management practice variables) along with physicochemical variables (Total Carbon, Total Nitrogen, and elemental (Al, As, B, Ca, Cd, Cr, Cu, Fe, K, Mg, Mn, Mo, Na, Ni, P, Pb, S, Si, Zn) composition) as dataset features to predict their impact on the MDR.

\subsection{Pathogen reduction in post-harvest pastured poultry practices}

This multi-classification investigation set includes features collected from poultry chicken samples during both the processing (Cecal) and postprocessing stages (whole carcass rinse (WCR) immediately following processing, as well as final product rinse (WCR) after chilling and storage). Food-borne pathogens \textit{Salmonella}, \textit{Listeria}, and \textit{Campylobacter} were identified and analyzed~\cite{rothrock2016antibiotic} A number of common poultry farm practices, as well as processing, water, freezing, and storage practices, are analyzed to identify the source of food-borne diseases.



\begin{figure*}[t]
\centering
\begin{subfigure}[t]{0.85\textwidth}

\includegraphics[width=0.9\textwidth]{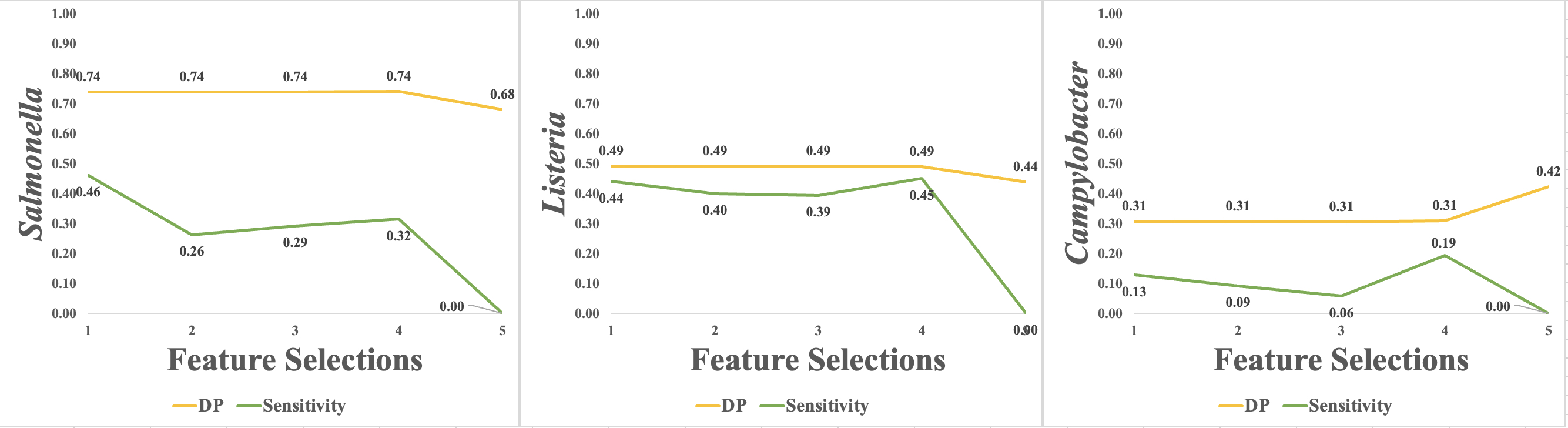}
\caption{Pathogen prediction results when tested on post-harvest poultry farm data.}
\label{fig:post_results}
\end{subfigure}
\begin{subfigure}[t]{0.85\textwidth}

\includegraphics[width=0.9\textwidth]{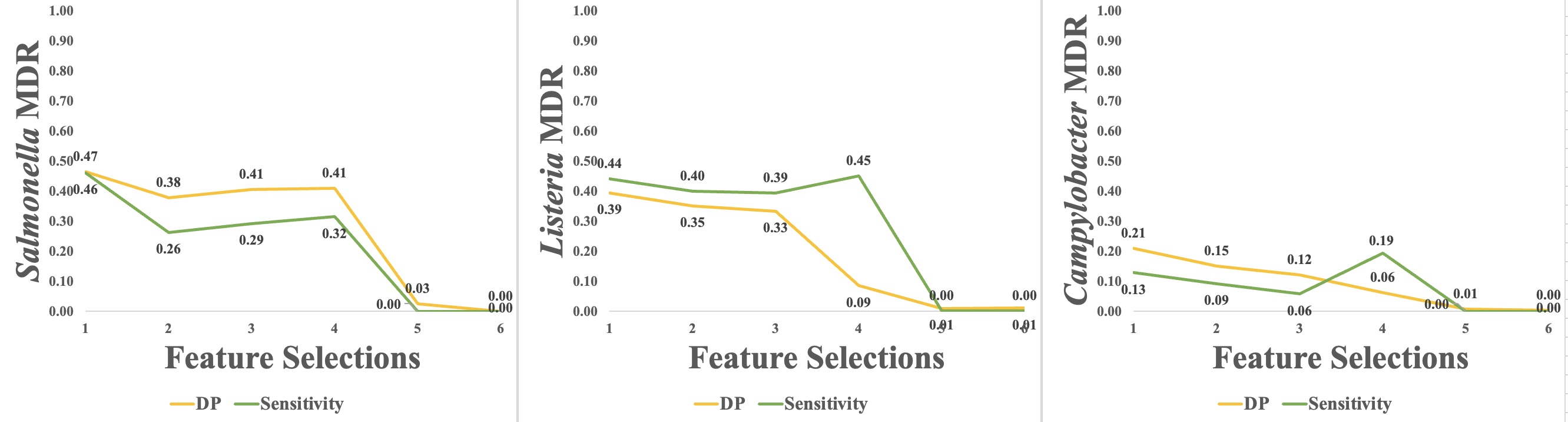}
\caption{Multi-drug resistance prediction results when tested on pre-harvest poultry farm data.}
\label{fig:pre_results}
\end{subfigure}
\caption{Comparison of pathogen prediction while finding optimum combinatorial optimization based on explanation-based approach (sensitivity) and dynamic programming (DP). Feature selection is carried out sequentially, combining those that are most influential in reducing pathogen rates. These results indicate that explanation-based approaches are capable of identifying combinations of features that reduce pathogen presence in fewer iterations than DP.} 
\label{fig:results}
\end{figure*}

\section{Experiments And Results} \label{experiments}

Our experiments used a hidden size of 30 when building the initial neural network model and 100 samples as a reference for SHAP. The pruning threshold is used to eliminate any data around zero to reduce the search space. The results were confirmed by four rounds of experiments. To verify the efficiency of our models, we tested them on two datasets. We used the ReLU function as a
non-linear activation function in the hidden layers, followed by
a sigmoid layer as it is a classification problem. Our models were implemented using tensor-flow, keras, scikit-learn, and Python.

\noindent \textbf{Baseline:} We implemented a dynamic programming (DP) optimization algorithm~\cite{Pillai_CO2023} as a baseline to evaluate the efficiency of our combinatorial optimization model for multi-label classification. It operates by sequentially computing optimal predictions for each feature subset and selecting the best combination based on the full set of results. We assessed both this traditional DP architecture and our proposed architecture on two multi-label classification datasets. By benchmarking against dynamic programming, we aimed to verify whether our model can deliver comparable performance to this established optimization strategy through more efficient, heuristic methodology. The two architectures were rigorously validated and contrasted using dataset classifications.

In figures \ref{fig:post_results}, \ref{fig:pre_results}, the pathogen prediction for explanation-based method compared to dynamic programming is shown at each combination of feature selections. With the right combination of farm management practices, we could see a reduction in the prevalence of the pathogens. As can be seen in the figure~\ref{fig:post_results}, the pathogen levels have decreased rapidly in the pre-harvest dataset compared to DP. In figure~\ref{fig:pre_results}, we can see the reduction in multi-drug resistance in the post-harvest dataset. A challenging part of reducing Listeria MDR is finding the best combinations of farm practices due to the imbalance between positives and negatives, which is where dynamic programming is useful.

\begin{figure*}[t]
\centering
\includegraphics[width=0.8\textwidth]{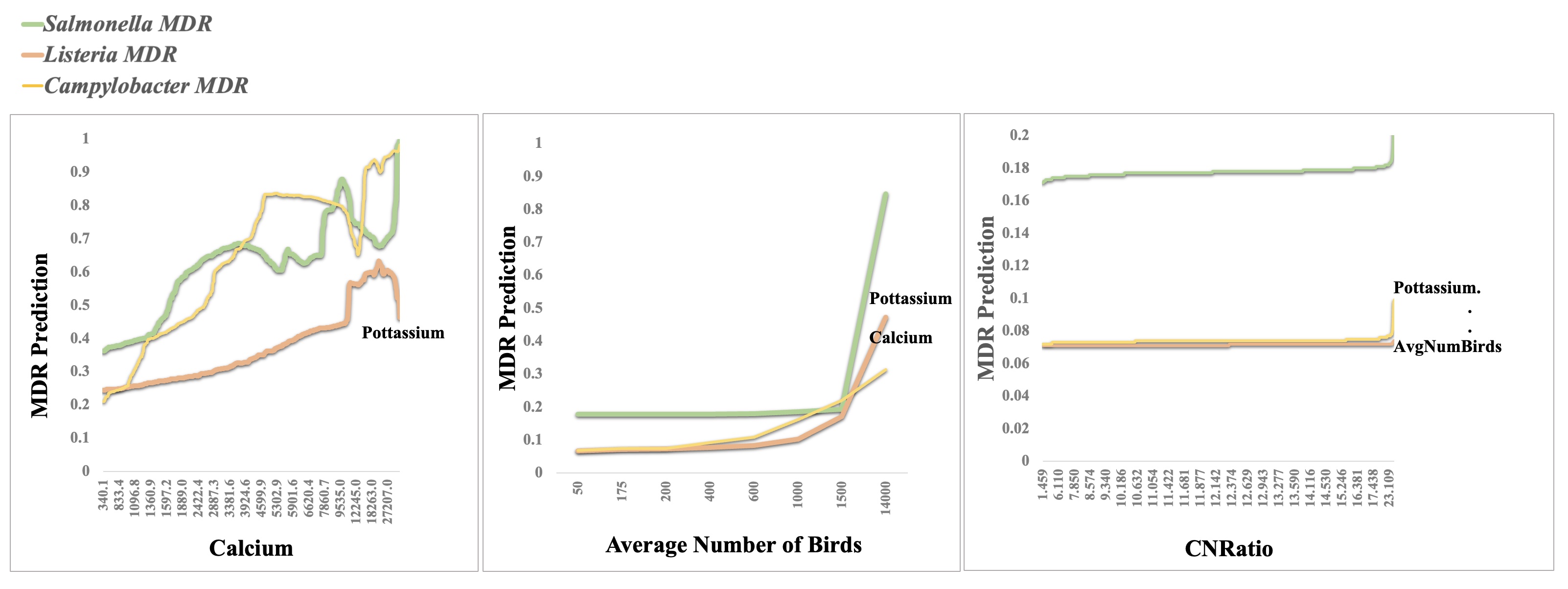}

\caption{Multi-drug resistance (MDR) predictions of a single variable (x-axis label) in the pre-harvest dataset in combination with other features (right middle labels in all figures). We select the feature value for these variables that produces the lowest average MDR prediction score across the three pathogens modeled. By considering all three pathogens simultaneously, the optimization accounts for interactions between conditions favoring differing types of pathogen growth.}
\label{fig:preharvest_results_feature_interaction}
\end{figure*}

\begin{figure*}[t]
\centering
\includegraphics[width=0.8\textwidth]{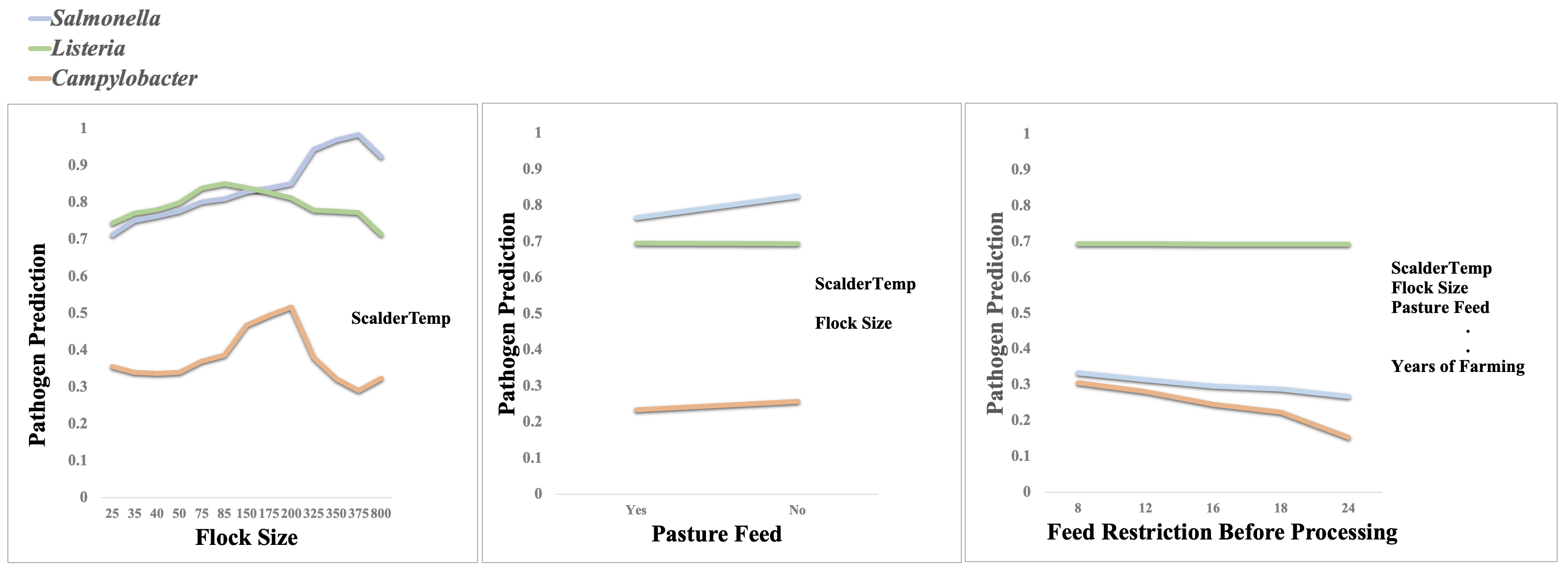}


\caption{Pathogen predictions of a single variable (x-axis label) in the post-harvest dataset in combination with other features (right middle labels in all figures). We select the feature value for these variables that produces the lowest average pathogen prediction score across the three pathogens modeled. By considering all three pathogens simultaneously, the optimization accounts for interactions between conditions favoring differing types of pathogen growth.}
\label{fig:postharvest_results_feature_interaction}
\end{figure*}

Figures \ref{fig:preharvest_results_feature_interaction} and \ref{fig:postharvest_results_feature_interaction} illustrate how a single variable (x-axis label) influences pathogen prediction (y-axis label) when combined with other variables (middle labels in all figures). The feature value for each of these variables is selected that produces the lowest average pathogen prediction score across all three pathogen models. In the sections \ref{result_pre} and \ref{result_post}, we detail the feature combinations that were found to be highly influential in reducing pathogen presence.

\subsection{Multi-drug resistance reduction in pre-harvest pastured poultry practices} \label{result_pre}

\paragraph{Potassium (K)}  Increasing potassium in soils can encourage overall healthy microbial life in the soil. This diverse microbial community can outcompete pathogens and prevent them from establishing strong populations. In our evaluations, we discovered that high levels of potassium in soil samples correlated with lower amounts of all three pathogens. When the potassium concentrations decreased, the prevalence of these poultry pathogens tended to increase.

\paragraph{Calcium (Ca)} Calcium supports survival and virulence of certain \textit{Salmonella} strains. By limiting dietary calcium, the pathogens may become less viable and infectious~\cite{ricke2004feeding}. The combination of low calcium levels with high potassium levels also contributed to the reduction of MDR scores in our experiment. 

\paragraph{Average Number of Birds} Farms with fewer birds can more easily monitor and implement safety measures for each individual chicken, which may indirectly limit new pathogens from entering the flock and spreading among birds. Based on results from soil and feces samples, it has consistently been shown that limiting the number of birds (less than 500) in a poultry farm is associated with a reduction in multi-drug resistance.
\paragraph{Carbon-to-Nitrogen Ratio (CNRatio)} The balance between carbon and nitrogen in poultry production systems may help mitigate antibiotic-resistant pathogens. CNRatio, however, did not appear to have a significant effect on multi-drug resistance levels in our pre-harvest samples.

\paragraph{Antibiotics Usage} Detection of antibiotic resistant bacteria in farms that do not use antibiotics routinely tends to be lower than that in farms that use antibiotics heavily~\cite{shrestha2022associations}. Our study results are in agreement with these findings, as no antibiotics were used in farms, suggesting a lower rate of multi-drug resistance. 

\subsection{Pathogen reduction in post-harvest pastured poultry practices}\label{result_post}

\paragraph{Scalder Temperature} In the study, scalder water temperature during processing was identified as one of the most impactful practices for reducing pathogen contamination levels on finished chicken carcasses. Heat helps enable microbial reductions on carcasses, but extreme temperatures cause cell damage. Correct scalder temperature range is the key - too hot risks pushing pathogens into skin crevices while too cold leaves organic materials for microbes. Our tests revealed that scalder temperatures of 20 degrees Celsius were the most conducive to reducing pathogens.

\paragraph{FlockSize}
In larger flocks, \textit{Salmonella} thrives since there are more routes for the bacteria to spread. More birds shedding bacteria amplify the spread of the bacteria through contaminated dust, litter, workers, equipment, etc. In both very low and very high flock sizes, \textit{Listeria} and \textit{Campylobacter} prevalence declines for different reasons. There are stricter biosecurity measures, less exposure and more strenuous oversight in small flocks, whereas there are commercial countermeasures, automated monitoring and rapid tracing in large flocks. However, medium-sized flocks are not protected in either extreme.

\paragraph{Pasture Feed}
As evidenced in our findings, providing pasture access and foraged feed is a positive contributing factor for reducing pathogen levels in poultry production systems.

\paragraph{Brood Cleaning Frequency} 

According to Hwang~\cite{hwang2020farm}, to reduce the prevalence of \textit{Salmonella}, equipment should be cleaned and sanitized regularly after each flock. The improvement of brood housing hygiene practices directly contributes to the eradication of infectious pathogens between grow outs and the emptying of the housing between grow outs. Our results are consistent with this and demonstrate that cleaning should be performed on a weekly basis, at least once a year in the case of large farms, to limit the spread of pathogens.

\paragraph{Soy Free Brood Food}
Feeding chickens soy-free diets can help avoid gut irritation from soy proteins. Our results found that raising chickens without soy feed produces higher quality chicken meat with fewer pathogens as compared with raising chickens on soy feed.

\paragraph{Flock Age in Days}
Based on our findings at 55 days compared to up to 125 days, the lower the flock age, the lower the pathogen levels. As the flock ages, opportunities for exposure build up over time via contaminated water, feed, workers, pests, equipment etc. Immune function and gut integrity in broilers  declines as the chickens near market weight. This makes older flocks more susceptible. However, in experiments conducted with samples containing whole carcass rinse immediately after processing (WCR-P), \textit{Campylobacter} isolation was lower in flocks older than 10 weeks~\cite{xu2021using}.

\paragraph{Water Source}
Our findings suggest that using rain water as the source for farms may have some advantages over well water when it comes to reducing potential pathogen contamination. The quality of rainwater is not impacted by groundwater contamination, as is the case with well water. The use of water from wells for rinsing after processing is not recommended.

\paragraph{Years of Farming}
The fewer years of farming (2 years) likely contributed to the low pathogen rates for all three pathogens. Hwang et.al \cite{hwang2020farm} report shows a similar prediction of \textit{Salmonella} prevalence based on post-harvest data. Long-term exposure over decades of farming may indicate that various pathogens have been encountered on the farm. It is possible for pathogens to become more resistant to treatment programs that were previously effective.

\paragraph{Length of Feed Restriction before Processing}

We tested 8 to 24 hours of feed restriction for reducing pathogen levels, and 24 hours of feed restriction seemed to be the most effective for reducing pathogen levels. The fasting period of 24 hours helps clear the digestive tract and reduces pathogen contamination during the processing phase. The management practice (perhaps extended feed restriction before slaughter) is reasonable when evaluating post-harvest pathogen data during processing, but may not appear applicable to pre-harvest pathogen data.

\paragraph{Average Number of Birds}
According to our findings, poultry farms with fewer bird populations, typically 200 or fewer, when compared with poultry farms with 50 to 14,000 birds, showed lower pathogen levels. Farmers have fewer birds to manage, which makes it easier for them to implement tight biosecurity protocols, monitor each bird closely for illness, quarantine sick birds, and clean meticulously between harvests. It should be noted, however, that even large commercial systems have advantages for comprehensive disease control through automated environmental systems, specialized staff, and traceback mechanisms. An ideal number can be difficult to suggest because a variety of factors need to be considered, including production type, housing dimensions, and surveillance capabilities.

\subsection{Scope \& Limitation} \label{scope}

We have demonstrated our approach for identifying combinations of features to achieve specific goals in pastured poultry datasets. However, this approach holds value in any domain where optimizing combinations of features for particular needs is essential. In this section, we describe the application of our approach in other domains and its scalability for larger datasets.

Agricultural scientists can use combinatorial optimization to search for needles within haystacks - whether it is for crop plans, resource plans, breeding strategies, or monitoring infrastructure. Our approach can contribute to generating and evaluating an enormous number of crop rotation sequences in order to maximize yields, minimize soil nutrient depletion, and conform to various constraints. In addition, we can use our approach to develop algorithms for scheduling agricultural robots and drones, scheduling algorithms for automated irrigation systems, and sub-problems in precision agriculture based on combinatorial optimization. The technique may also be employed to reconstruct the most likely genetic sequence from overlapping DNA segments by solving complex combinatorial puzzles. The use of combinatorial optimization is also of great value in solving many core problems in computer science, engineering, and operations research. These include parameter optimization for neural networks and machine learning models, optimization of vehicle routing and warehousing, optimization of network topology and flow routing, and optimization of parts placements to minimize production costs.

Combinatorial optimization algorithms can face significant scalability and computational efficiency challenges when applied to larger, more complex real-world datasets and problem instances. The number of possible solutions often grows exponentially with problem size. However, our pruning approach reduces the complexity from growing exponentially to linearly. A hierarchical and multi-stage decomposition of large problems is recommended for exact algorithms, which we implement using our two-stage methodology.

\section{Conclusion} \label{conclusion}
The purpose of this study is to identify patterns/combinations that can predict a desired phenotype from agricultural data. Utilizing explanation-based combinatorial optimization, which supports multi-objective classification, we propose a combinatorial selection procedure for determining which farm management practices reduce pathogen prevalence most effectively. The results demonstrate that this explainable based combinatorial approach, combined with a classification model, can be useful for reducing the spread of pathogens in pastured poultry management systems.

\section{Funding} \label{funding}

Dataset used in this study is provided by the Agricultural Research Service, USDA CRIS Project ``Reduction of Invasive \textit{Salmonella enterica} in Poultry through Genomics, Phenomics and Field Investigations of Small MultiSpecies Farm Environments'' \#6040-32000-011-00-D. This research was supported by the Agricultural Research Service, USDA NACA project entitled ``Advancing Agricultural Research through High Performance Computing'' \#58-0200-0-002.


\bibliography{reference}

\begin{thebibliography}{10}
\providecommand{\url}[1]{#1}
\csname url@samestyle\endcsname
\providecommand{\newblock}{\relax}
\providecommand{\bibinfo}[2]{#2}
\providecommand{\BIBentrySTDinterwordspacing}{\spaceskip=0pt\relax}
\providecommand{\BIBentryALTinterwordstretchfactor}{4}
\providecommand{\BIBentryALTinterwordspacing}{\spaceskip=\fontdimen2\font plus
\BIBentryALTinterwordstretchfactor\fontdimen3\font minus \fontdimen4\font\relax}
\providecommand{\BIBforeignlanguage}[2]{{%
\expandafter\ifx\csname l@#1\endcsname\relax
\typeout{** WARNING: IEEEtranS.bst: No hyphenation pattern has been}%
\typeout{** loaded for the language `#1'. Using the pattern for}%
\typeout{** the default language instead.}%
\else
\language=\csname l@#1\endcsname
\fi
#2}}
\providecommand{\BIBdecl}{\relax}
\BIBdecl

\bibitem{acampora2023genetic}
G.~Acampora, A.~Chiatto, and A.~Vitiello, ``Genetic algorithms as classical optimizer for the quantum approximate optimization algorithm,'' \emph{Applied Soft Computing}, vol. 142, p. 110296, 2023.

\bibitem{ahmed2022artificial}
I.~Ahmed, G.~Jeon, and F.~Piccialli, ``From artificial intelligence to explainable artificial intelligence in industry 4.0: a survey on what, how, and where,'' \emph{IEEE Transactions on Industrial Informatics}, vol.~18, no.~8, pp. 5031--5042, 2022.

\bibitem{ayoola2022preharvest}
M.~B. Ayoola, N.~Pillai, B.~Nanduri, M.~J. Rothrock, and M.~Ramkumar, ``Preharvest environmental and management drivers of multidrug resistance in major bacterial zoonotic pathogens in pastured poultry flocks,'' \emph{Microorganisms}, vol.~10, no.~9, p. 1703, 2022.

\bibitem{azzini2021sobol}
I.~Azzini and R.~Rosati, ``Sobol’main effect index: an innovative algorithm (ia) using dynamic adaptive variances,'' \emph{Reliability Engineering \& System Safety}, vol. 213, p. 107647, 2021.

\bibitem{de2022random}
M.~B. De~Moraes and G.~P. Coelho, ``A random forest-assisted decomposition-based evolutionary algorithm for multi-objective combinatorial optimization problems,'' in \emph{2022 IEEE Congress on Evolutionary Computation (CEC)}.\hskip 1em plus 0.5em minus 0.4em\relax IEEE, 2022, pp. 1--8.

\bibitem{geist2023combinatorial}
E.~L. Geist and T.~Parsons, ``Combinatorial optimization of earthquake spatial distributions under minimum cumulative stress constraints,'' \emph{Bulletin of the Seismological Society of America}, 2023.

\bibitem{gunning2019xai}
D.~Gunning, M.~Stefik, J.~Choi, T.~Miller, S.~Stumpf, and G.-Z. Yang, ``Xai—explainable artificial intelligence,'' \emph{Science robotics}, vol.~4, no.~37, p. eaay7120, 2019.

\bibitem{hurtado2022exact}
O.~G. Hurtado, R.~P. Ch, and J.~Moncada, ``Exact and approximate sequential methods in solving the quadratic assignment problem,'' \emph{Journal of Language and Linguistic Studies}, vol.~18, no.~3, 2022.

\bibitem{hwang2020farm}
D.~Hwang, M.~J. Rothrock~Jr, H.~Pang, G.~D. Kumar, and A.~Mishra, ``Farm management practices that affect the prevalence of salmonella in pastured poultry farms,'' \emph{Lwt}, vol. 127, p. 109423, 2020.

\bibitem{kaya2022review}
E.~Kaya, B.~Gorkemli, B.~Akay, and D.~Karaboga, ``A review on the studies employing artificial bee colony algorithm to solve combinatorial optimization problems,'' \emph{Engineering Applications of Artificial Intelligence}, vol. 115, p. 105311, 2022.

\bibitem{laato2022explain}
S.~Laato, M.~Tiainen, A.~Najmul~Islam, and M.~M{\"a}ntym{\"a}ki, ``How to explain ai systems to end users: a systematic literature review and research agenda,'' \emph{Internet Research}, vol.~32, no.~7, pp. 1--31, 2022.

\bibitem{lundberg2017unified}
S.~M. Lundberg and S.-I. Lee, ``A unified approach to interpreting model predictions,'' \emph{Advances in neural information processing systems}, vol.~30, 2017.

\bibitem{NIPS2017_7062}
\BIBentryALTinterwordspacing
------, ``A unified approach to interpreting model predictions,'' in \emph{Advances in Neural Information Processing Systems 30}, I.~Guyon, U.~V. Luxburg, S.~Bengio, H.~Wallach, R.~Fergus, S.~Vishwanathan, and R.~Garnett, Eds.\hskip 1em plus 0.5em minus 0.4em\relax Curran Associates, Inc., 2017, pp. 4765--4774. [Online]. Available: \url{http://papers.nips.cc/paper/7062-a-unified-approach-to-interpreting-model-predictions.pdf}
\BIBentrySTDinterwordspacing

\bibitem{majumder2022multi}
S.~Majumder, P.~S. Barma, A.~Biswas, P.~Banerjee, B.~K. Mandal, S.~Kar, and P.~Ziemba, ``On multi-objective minimum spanning tree problem under uncertain paradigm,'' \emph{Symmetry}, vol.~14, no.~1, p. 106, 2022.

\bibitem{panigutti2022understanding}
C.~Panigutti, A.~Beretta, F.~Giannotti, and D.~Pedreschi, ``Understanding the impact of explanations on advice-taking: a user study for ai-based clinical decision support systems,'' in \emph{Proceedings of the 2022 CHI Conference on Human Factors in Computing Systems}, 2022, pp. 1--9.

\bibitem{panwar2023discrete}
K.~Panwar and K.~Deep, ``Discrete salp swarm algorithm for euclidean travelling salesman problem,'' \emph{Applied Intelligence}, vol.~53, no.~10, pp. 11\,420--11\,438, 2023.

\bibitem{pillai2022ensemble}
N.~Pillai, M.~B. Ayoola, B.~Nanduri, M.~J. Rothrock~Jr, and M.~Ramkumar, ``An ensemble learning approach to identify pastured poultry farm practice variables and soil constituents that promote salmonella prevalence,'' \emph{Heliyon}, vol.~8, no.~11, p. e11331, 2022.

\bibitem{Pillai_CO2023}
N.~Pillai, B.~Nanduri, M.~J. Rothrock, Z.~Chen, and M.~Ramkumar, ``Towards optimal microbiome to inhibit multidrug resistance,'' in \emph{2023 IEEE Conference on Computational Intelligence in Bioinformatics and Computational Biology (CIBCB)}, 2023, pp. 1--9.

\bibitem{ricke2004feeding}
S.~Ricke, S.~Park, R.~Moore, Y.~Kwon, C.~Woodward, J.~Byrd, D.~Nisbet, and L.~Kubena, ``Feeding low calcium and zinc molt diets sustains gastrointestinal fermentation and limits salmonella enterica serovar enteritidis colonization in laying hens,'' \emph{Journal of food safety}, vol.~24, no.~4, pp. 291--308, 2004.

\bibitem{rothrock2016antibiotic}
M.~J. Rothrock~Jr, K.~L. Hiett, J.~Y. Guard, and C.~R. Jackson, ``Antibiotic resistance patterns of major zoonotic pathogens from all-natural, antibiotic-free, pasture-raised broiler flocks in the southeastern united states,'' \emph{Journal of Environmental Quality}, no.~2, pp. 593--603, 2016.

\bibitem{shrestha2022associations}
R.~D. Shrestha, A.~Agunos, S.~P. Gow, A.~E. Deckert, and C.~Varga, ``Associations between antimicrobial resistance in fecal escherichia coli isolates and antimicrobial use in canadian turkey flocks,'' \emph{Frontiers in Microbiology}, vol.~13, p. 954123, 2022.

\bibitem{shrikumar2017learning}
A.~Shrikumar, P.~Greenside, and A.~Kundaje, ``Learning important features through propagating activation differences,'' in \emph{International conference on machine learning}.\hskip 1em plus 0.5em minus 0.4em\relax PMLR, 2017, pp. 3145--3153.

\bibitem{shrikumar2016not}
A.~Shrikumar, P.~Greenside, A.~Shcherbina, and A.~Kundaje, ``Not just a black box: Learning important features through propagating activation differences,'' \emph{arXiv preprint arXiv:1605.01713}, 2016.

\bibitem{tahami2022literature}
H.~Tahami and H.~Fakhravar, ``A literature review on combining heuristics and exact algorithms in combinatorial optimization,'' \emph{European Journal of Information Technologies and Computer Science}, vol.~2, no.~2, pp. 6--12, 2022.

\bibitem{weinand2022research}
J.~M. Weinand, K.~S{\"o}rensen, P.~San~Segundo, M.~Kleinebrahm, and R.~McKenna, ``Research trends in combinatorial optimization,'' \emph{International Transactions in Operational Research}, vol.~29, no.~2, pp. 667--705, 2022.

\bibitem{weissteiner2023bayesian}
J.~Weissteiner, J.~Heiss, J.~Siems, and S.~Seuken, ``Bayesian optimization-based combinatorial assignment,'' in \emph{Proceedings of the AAAI Conference on Artificial Intelligence}, vol.~37, no.~5, 2023, pp. 5858--5866.

\bibitem{xu2020deep}
S.~Xu, S.~S. Panwar, M.~Kodialam, and T.~Lakshman, ``Deep neural network approximated dynamic programming for combinatorial optimization,'' in \emph{Proceedings of the AAAI Conference on Artificial Intelligence}, vol.~34, no.~02, 2020, pp. 1684--1691.

\bibitem{xu2021using}
X.~Xu, M.~J. Rothrock~Jr, A.~Mohan, G.~D. Kumar, and A.~Mishra, ``Using farm management practices to predict campylobacter prevalence in pastured poultry farms,'' \emph{Poultry Science}, vol. 100, no.~6, p. 101122, 2021.

\bibitem{yang2023memory}
H.~Yang, M.~Zhao, L.~Yuan, Y.~Yu, Z.~Li, and M.~Gu, ``Memory-efficient transformer-based network model for traveling salesman problem,'' \emph{Neural Networks}, vol. 161, pp. 589--597, 2023.

\bibitem{zhang2023survey}
J.~Zhang, C.~Liu, X.~Li, H.-L. Zhen, M.~Yuan, Y.~Li, and J.~Yan, ``A survey for solving mixed integer programming via machine learning,'' \emph{Neurocomputing}, vol. 519, pp. 205--217, 2023.

\bibitem{zhang2022deep}
T.~Zhang, A.~Banitalebi-Dehkordi, and Y.~Zhang, ``Deep reinforcement learning for exact combinatorial optimization: Learning to branch,'' in \emph{2022 26th International Conference on Pattern Recognition (ICPR)}.\hskip 1em plus 0.5em minus 0.4em\relax IEEE, 2022, pp. 3105--3111.

\bibitem{zhao2022evolution}
B.~Zhao, W.-N. Chen, F.-F. Wei, X.~Liu, Q.~Pei, and J.~Zhang, ``Evolution as a service: A privacy-preserving genetic algorithm for combinatorial optimization,'' \emph{arXiv preprint arXiv:2205.13948}, 2022.

\bibitem{zhu2022improved}
D.~Zhu, Z.~Huang, L.~Xie, and C.~Zhou, ``Improved particle swarm based on elastic collision for dna coding optimization design,'' \emph{IEEE Access}, vol.~10, pp. 63\,592--63\,605, 2022.

\end{thebibliography}

\end{document}